\documentclass[letterpaper]{article} 
\usepackage{aaai2026}  
\usepackage{times}  
\usepackage{helvet}  
\usepackage{courier}  
\usepackage[hyphens]{url}  
\usepackage{graphicx} 
\usepackage{natbib}  
\usepackage{caption} 
\usepackage{algorithm}
\usepackage{algorithmic}

\usepackage{newfloat}
\usepackage{listings}

\usepackage{booktabs}
\usepackage{amsmath}
\usepackage{amssymb}
\usepackage{multirow}
\usepackage{placeins}

\DeclareCaptionStyle{ruled}{labelfont=normalfont,labelsep=colon,strut=off} 
\floatstyle{ruled}
\newfloat{listing}{tb}{lst}{}
\floatname{listing}{Listing}
\title{4DSTR: Advancing Generative 4D Gaussians with Spatial-Temporal Rectification for High-Quality and Consistent 4D Generation}
\author{
    Mengmeng Liu\textsuperscript{\rm 1}\equalcontrib, Jiuming Liu\textsuperscript{\rm 2}\equalcontrib, Yunpeng Zhang\textsuperscript{\rm 3},  Jiangtao Li\textsuperscript{\rm 3},\\
    Michael Ying Yang\textsuperscript{\rm 4}, Francesco Nex\textsuperscript{\rm 1}, Hao Cheng\textsuperscript{\rm 1}\thanks{Corresponding author.}
}
\affiliations{
    \textsuperscript{\rm 1} University of Twente \textsuperscript{\rm 2} University of Cambridge \textsuperscript{\rm 3} PhiGent Robotics \textsuperscript{\rm 4} University of Bath \\
    m.liu-1@utwente.nl, jl2538@cam.ac.uk, h.cheng-2@utwente.nl

}

\begin{document}




\maketitle

\begin{figure*}
    
\includegraphics[width=\textwidth]{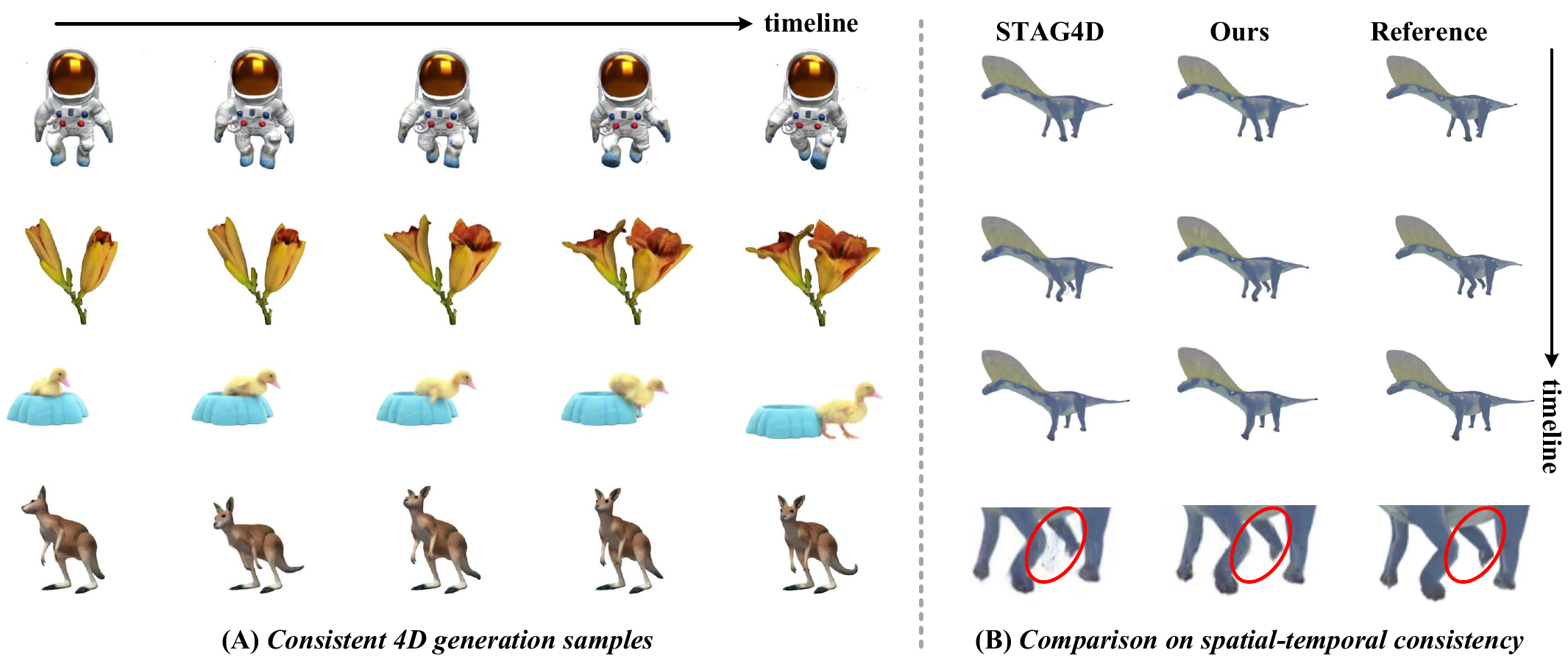}
\captionof{figure}{%
  \textbf{Consistent 4D generation with spatial‐temporal rectification.}
  Our method proposes a novel framework for high‐quality 4D generation as in (A).
  Compared to the state‐of‐the‐art method STAG4D \cite{zeng2024stag4d}, our method has higher generation
  consistency and quality in the dynamic region (red circle) of generated 4D sequences (B), which demonstrates
  that our rectification methods significantly boost spatial‐temporal consistency in generative 4D Gaussian
  representations.
}
\label{fig:vis}
\end{figure*}

\begin{abstract}

Remarkable advances in recent 2D image and 3D shape generation have induced a significant focus on dynamic 4D content generation. However, previous 4D generation methods commonly struggle to maintain spatial-temporal consistency and adapt poorly to rapid temporal variations, due to the lack of effective spatial-temporal modeling. To address these problems, we propose a novel 4D generation network called 4DSTR, which modulates generative 4D Gaussian Splatting with spatial-temporal rectification. Specifically, temporal correlation across generated 4D sequences is designed to rectify deformable scales and rotations and guarantee temporal consistency. Furthermore, an adaptive spatial densification and pruning strategy is proposed to address significant temporal variations by dynamically adding or deleting Gaussian points with the awareness of their pre-frame movements. Extensive experiments demonstrate that our 4DSTR achieves state-of-the-art performance in video-to-4D generation, excelling in reconstruction quality, spatial-temporal consistency, and adaptation to rapid temporal movements.
\end{abstract}


\section{Introduction}
\label{sec:intro}
Recently, there have been advancements in generating high-quality and diverse visual contents with pre-trained diffusion models \cite{ho2020denoising}, including 2D images \cite{ho2020denoising,zhou2025opening}, 3D shapes \cite{liu2025topolidm}, etc. These successful experiences naturally boost the exploration using generative diffusion models for dynamic 4D content generation \cite{singer2023text,bahmani2024tc4d,wu2024sc4d}, which has various applications in autonomous driving simulation \cite{liu2023translo,liu2024dvlo,cheng2023gatraj,cheng2023end,ni2025recondreamer}, virtual reality \cite{202508.0462,202508.0195}, and digital avatar animation \cite{wang2025humandreamer}, etc.

A common research line takes text \cite{singer2023text,ling2024align,wang20254real} as input conditions, leveraging pre-trained text-to-image or text-to-video diffusion models as the preprocess. MAV3D \cite{singer2023text} is a pioneering framework for text-to-4D generation that utilizes the pre-trained text-to-image and text-to-video diffusion models for the static and dynamic sequence generation, supervised by the Score Distillation Sampling (SDS) loss. Subsequent approaches design trajectory conditions \cite{bahmani2024tc4d} or deformable 4D Gaussian Splatting \cite{ling2024align} to enhance the motion fidelity and appearance quality of generated 4D samples. However, recent text-to-4D generation networks fail to capture the multi-view spatial-temporal consistency and struggle with effective knowledge distillation from diffusion models as in Fig. \ref{fig:vis}.

Another research line focuses on the video-to-4D generation task. Consistent4D \cite{jiangconsistent4d} firstly proposes an Interpolation-driven Consistency Loss to enforce the similarity between reconstruction samples from DyNeRF \cite{fridovich2023k} and interpolated frames. However, the implicit nature of neural representation leads to long optimization time and poor motion controllability. With the great progress of explicit 4D Gaussian Splatting \cite{wu20244d}, more researchers \cite{ren2023dreamgaussian4d,wu2024sc4d,li2024dreammesh4d,wu2025cat4d} formulate the deformable Gaussian Splatting as the intermediate 4D representation. However, these video-to-4D generation pipelines still encounter challenges like spatial-temporal inconsistency and suboptimal motion quality. STAG4D \cite{zeng2024stag4d} designs temporal anchor points for enhanced 4D consistency, but lacks temporal correlation, and their designed Gaussian densification technique fails to consider rapid texture differences in the same region across frames, as shown in Fig. \ref{fig:adaptive1}, and overlooks the time-varying requirements in the number of Gaussian points for adaptive appearance modeling. 

To address these challenges, we propose a novel 4D generation framework called \textbf{4DSTR}, which conducts the temporal correlation and spatial Gaussian rectification methods to guarantee spatial-temporal consistency and motion realism in generated 4D contents. Specifically, to ensure the temporal consistency of multi-frame generative 4D Gaussian points, the scales and rotations from all the video frames are correlated through the Mamba-based \cite{gu2023mamba,liu2024point} temporal encoding layer. Then the per-frame scale residuals and rotation residuals are generated to rectify the original inaccurate Gaussian points, which can significantly boost the inherent motion coherence of generated 4D sequences. Additionally, a spatial rectification method with the adaptive densification and pruning strategy is proposed to add Gaussian points to regions with rich textures requiring more 4D representation tokens and delete Gaussian points in regions with less texture or unstable Gaussian attributes for each training iteration. The adaptive densification and pruning strategy can enable the generative 4D Gaussian points to adapt to rapid temporal variations and possess more photorealistic reconstruction quality. Integrating these two designs, our method can significantly boost spatial-temporal consistency and generation quality for the video-to-4D generation task in Fig. \ref{fig:vis}. 
When combining with a pre-trained text-to-image or text-to-video generation diffusion model, our method can also support the text-to-4D generation task.

Overall, the contributions of this paper are as follows:
\begin{itemize}
    \item We propose 4DSTR, a novel 4D generation pipeline with spatial-temporal rectification to strengthen the spatial-temporal consistency of generated 4D videos and enhance the adaptation ability to rapid temporal variations.
    \item To ensure temporal consistency, a Mamba-based temporal encoding layer is designed to correlate video sequences and regress scale and rotation residuals of generative 4D Gaussian points in each frame.
    \item To adapt to rapid spatial variations across frames, we dynamically rectify the number of 4D Gaussian points using a per-frame adaptive Gaussian densification and pruning strategy with all-frame temporal awareness.
    \item Extensive experiments on the video-to-4D show the superiority of our proposed approach. Our 4DSTR outperforms the state-of-the-art 4D generation approach with a 15.1\% FID-VID reduction and a 19.9\% FVD reduction, which reveals our method's great potential in the spatial-temporal consistency of generated 4D sequences.
\end{itemize}

\section{Related Work}
\label{sec:related}

Recently, there has been a significantly increasing research focus on 4D generation with various guidance inputs including image, video \cite{jiangconsistent4d,ren2023dreamgaussian4d,wu2024sc4d,zeng2024stag4d}, and text \cite{singer2023text,ling2024align}. In this section, we mainly delve into descriptions for video-to-4D generation, text-to-4D generation, and also spatial-temporal modeling, respectively.

\subsection{Video-to-4D Generation}
Video-to-4D generation produces spatiotemporal content from uncalibrated monocular videos. Consistent4D \cite{jiangconsistent4d} applies an interpolation-driven loss on DyNeRF \cite{fridovich2023k} reconstructions but suffers from long optimization and limited motion control. DreamGaussian4D \cite{ren2023dreamgaussian4d} and SC4D \cite{wu2024sc4d} employ deformable 4D Gaussian splatting, while DreamMesh4D \cite{li2024dreammesh4d} combines mesh representations with geometric skinning for improved surface detail; however, they all lack strong spatial–temporal coherence. STAG4D \cite{zeng2024stag4d} introduces temporal anchors and adaptive densification to enhance frame-to-frame correlation. CAT4D \cite{wu2025cat4d} designs a sampling strategy to generate an unbounded collection of consistent multi-view videos for 4D generation. In contrast, we demonstrate that spatial-temporal modulation across frames is essential for enhanced consistency and high-quality motion.

\subsection{Text-to-4D Generation}
MAV3D \cite{singer2023text} pioneers by using a text-to-image diffusion model for static objects and a text-to-video model with SDS loss for dynamic sequences. AYG \cite{ling2024align} integrates compositional text-to-image, text-to-video, and 3D-aware multi-view diffusion to optimize 4D Gaussians, while TC4D \cite{bahmani2024tc4d} improves motion realism via trajectory-conditioned rigid transforms and local deformations. However, reliance on pre-trained diffusion models limits performance and leads to spatial–temporal inconsistency and domain gaps \cite{zhang20244diffusion}. 4Real-Video \cite{wang20254real} proposes a two-stream grid-based 4D video generation method with both viewpoint and temporal updates. 

\subsection{4D Spatial–Temporal Modeling}
4D spatial–temporal modeling extends static point cloud analysis to video by capturing temporal correlations \cite{liu2025dvlo4d,liu2023tracing,nie2025p2p,liu2024laformer,zhou2025focustrack,zhou2025pillarhist}. PSTNet \cite{fan2022pstnet} employs disentangled spatial and temporal convolutions, while P4Transformer \cite{fan2021point} uses Transformer for long-sequence embedding. Some researchers study scene flow methods \cite{liu2024difflow3d,jiang20243dsflabelling,jiang2024neurogauss4d,liu2023regformer} for 4D motion learning. Mamba4D \cite{liu2025mamba4d} uses the Mamba architecture \cite{gu2023mamba} to further enhance scalability. In this work, we transfer the successful experiences in 4D modeling into the spatial-temporal correlation of 4D Gaussian points for more consistent and high-quality 4D generation.

\section{Method}
\begin{figure*}[t!] 
\begin{center}

 \includegraphics[clip=true, trim=0in 0in 0in 0in, width=1\linewidth]{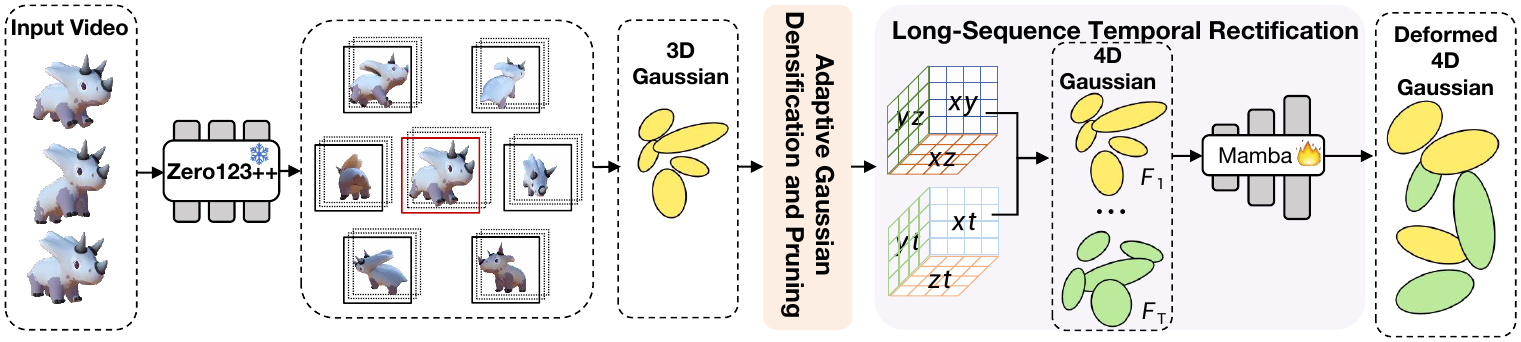} 
\end{center}
  \caption{{\textbf{The overall pipeline of our 4DSTR.}} Given an input video, we use Zero123++ \cite{shi2023zero123++} to generate multi-view frames and initialize the first-frame 3D Gaussians. A lightweight multi-head decoder then maps voxel features to per-frame 4D Gaussian parameters. To ensure 4D coherence, our temporal correlation module regresses scale and rotation residuals, while per-frame adaptive densification and pruning dynamically adjust Gaussian counts to capture rapid spatial changes.
}

\label{fig:pipeline}
\end{figure*}

In this section, we will dive into the details of our proposed 4D generation pipeline as illustrated in Fig. \ref{fig:pipeline}. First, we analyze the limitations of current 4D generative Gaussian representations. To mitigate these restrictions, we propose a temporal correlation module to establish temporal consistency and regress the scale and rotation residuals for each frame. Furthermore, an adaptive densification and pruning strategy is proposed to dynamically add or remove 4D Gaussian points at the frame level to strengthen the ability to adapt to rapid spatial variations over time.

\begin{figure}[t!] 
\begin{center}
 \includegraphics[clip=true, trim=0in 0in 0in 0in, width=0.9\linewidth]{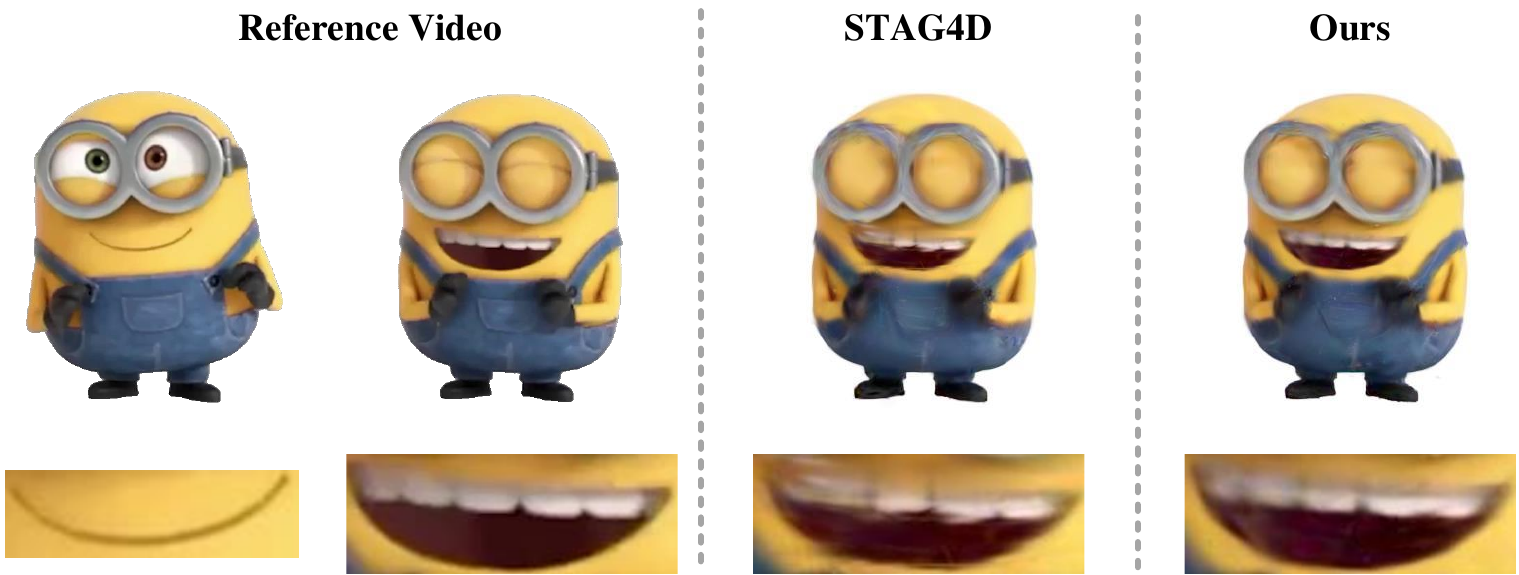} 
\end{center}
  \caption{{\textbf{Rapid temporal variations among frames.}} The mouth of Minions witnesses rapid appearance variations for two different frames. Compared to STAG4D \cite{zeng2024stag4d}, our method designs an adaptive Gaussian densification and pruning strategy, which largely enhances the adaptation capability of our 4D generative Gaussian.  }
\label{fig:adaptive1}
\end{figure}

\subsection{Limitations of Prior 4D Representations}
\label{subsec:4drepresentation}

Dynamic NeRF-based methods \cite{fridovich2023k} often suffer from poor motion consistency and flexibility due to their implicit nature and fixed bounding boxes \cite{bahmani2024tc4d}. Recent works \cite{zeng2024stag4d,ren2023dreamgaussian4d,li2024dreammesh4d} therefore adopt deformable 4D Gaussian Splatting \cite{wu20244d}, which extends 3D Gaussians \cite{kerbl20233d} with a deformation field:
\begin{equation}
\mathcal{F}(\mathcal{S},t) = [\mathcal{X}_t, s_t, r_t, \sigma, \zeta],
\end{equation}
where $\mathcal{X}_t$, $s_t$, $r_t$ update position, scale, and rotation, and $\sigma$, $\zeta$ correspond to the opacity and spherical harmonics (SH) coefficients of the radiance~\cite{kerbl20233d}, respectively.

However, these methods process each timestamp independently, lacking explicit temporal correlation or rectification \cite{zeng2024stag4d}. They also keep a constant number of Gaussians across frames, which hinders adaptation to rapid texture changes. As shown in Fig. \ref{fig:adaptive1}, suddenly changing regions with more texture details should be supplemented with more Gaussian points.  We address these issues by introducing a spatial-temporal correlation and rectification module alongside adaptive densification and pruning for enhanced spatial–temporal consistency and motion fidelity.

\subsection{Temporal Correlation and Rectification}
\label{sec:mem_out}
To effectively guarantee spatial-temporal consistency of generated 4D videos, we design a temporal buffer to store and facilitate interactions among multi-frame Gaussian attributes through the Mamba architecture \cite{gu2023mamba} as in Fig. \ref{fig:pipeline}. Unlike previous 4D Gaussian Splatting methods that only query Gaussian features from the current frame \cite{wu20244d}, our approach utilizes temporal Gaussian attributes ${\mathcal{F}}(\mathcal{S}, t) = [\mathcal{X}_t, s_t, r_t, \sigma, \zeta]_{t=1}^{T}$ from multiple timestamps as input for effective temporal correlation. Specifically, Gaussian attributes predicted from the current frame will correlate with the stored history Gaussian attributes in the temporal buffer to generate updated temporal features and regress the rectified Gaussian attribute residuals. The temporal buffer \(M^G_0 \in \mathbb{R}^{T\times d_G}\), which holds Gaussian attributes over time, is initialized empty and retains a fixed length \(T\). Here, \(d_G=11\) corresponds to the Gaussian attribute vectors: position, scale, rotation, and opacity.

\noindent\textbf{Temporal Correlation with Mamba.} During the temporal correlation phase, the current Gaussian attribute feature ${\mathcal{F}}_{t}$ just generated from the deformable network will interact with the temporal buffer through a sliding window mechanism. When current Gaussian attributes are written into the temporal buffer $M^\mathrm{G}_{t-1}$, the temporally farthest Gaussian attribute $\hat{\mathcal{F}}_{t-T-1}$ is discarded. The $T-1$ most recent Gaussian attributes $\{\hat{\mathcal{F}}_{t-T}, ..., \hat{\mathcal{F}}_{t-1}\}$ are then concatenated with the current Gaussian attribute $F_{t}$ as follows:
\begin{equation}
    \label{eq:concat}
    \{\hat{\mathcal{F}}_{t-T-1}, \hat{\mathcal{F}}_{t-T}, ..., \hat{\mathcal{F}}_{t-1}\} \Rightarrow \{\hat{\mathcal{F}}_{t-T}, ..., \hat{\mathcal{F}}_{t-1}, {\mathcal{F}}_{t}\},
\end{equation}
where the braces denote the concatenation of Gaussian attributes along the temporal axis. Concatenated features $M^\mathrm{G}_{t-1} = \{\hat{\mathcal{F}}_{t-T}, ..., \hat{\mathcal{F}}_{t-1}, {\mathcal{F}}_{t}\}$ are then passed as input tokens into the subsequent Mamba-based temporal correlation module:
\begin{equation}
    \label{eq:mamba1}
    \hat{\mathcal{F}}_{t} = \text{Mamba}(M^\mathrm{G}_{t-1}),
\end{equation}
where Mamba refers to the standard selective state-space model \cite{gu2023mamba}. The temporally-correlated Gaussian attributes are then used to regress the scale and rotation residuals for the rectification.

\noindent\textbf{Gaussian Attribute Rectification.} After the temporal correlation, we rectify the scales and rotations of Gaussian points for each frame. Here, we only extend details of the scale rectification, and the same procedure is applied to rotation modulation. We rectify the Gaussian scales in the current frame through the feature fusion of temporally-correlated Gaussian attribute features $\hat{\mathcal{F}}_{t}$,  current scales $s_{t}$, and history scales $\hat{s}_{t-1}$ as:
\begin{equation}
    \label{eq:feature gathering_2}
     \Delta s_{t} = W(\hat{\mathcal{F}}_{t} \oplus s_{t} \oplus \hat{s}_{t-1}),
\end{equation}
where $W$ means the dynamic weighting method in \cite{aydemir2023adapt}. $\hat{s}_{t-1}$ indicates the nearest history scales, and $\Delta s_{t}$ represents the regressed residuals of Gaussian scales. Then the rectified scales in the timestamp $t$ can be represented by: $\hat{s}_{t} = s_{t} + \Delta s_{t}$. Finally, we update the temporal buffers by concatenating the current temporally-correlated features \( \hat{\mathcal{F}}_t \) with the previous \( T-1 \) frames:
\begin{equation}
    \label{eq:concat2}
     M^\mathrm{G}_\mathrm{t} = \{\hat{\mathcal{F}}_\text{t-T+1}, ..., \hat{\mathcal{F}}_\text{t-1}, \hat{\mathcal{F}}_\text{t}\}.
\end{equation}
$M^\mathrm{G}_\mathrm{t}$ will be used for the rectification process in the next timestamp $t+1$, regressing scale and rotation residuals $\Delta s_{t+1}$ and $\Delta r_{t+1}$.

\subsection{Adaptive Gaussian Densification and Pruning}
\label{sec:spatial}

\begin{figure}[t!] 
\begin{center}

 \includegraphics[clip=true, trim=0in 0in 0in 0in, width=0.75\linewidth]{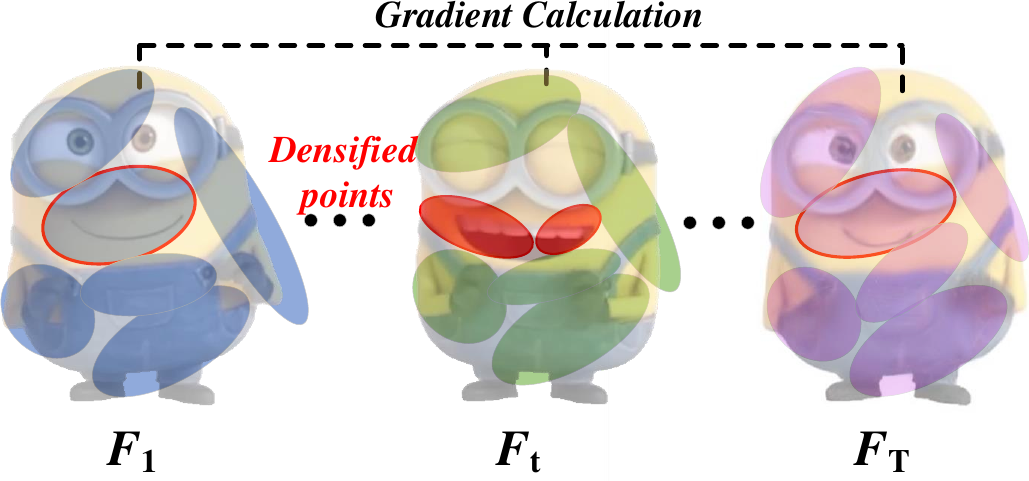} 
\end{center}
  \caption{{\textbf{Illustration of Per-frame Adaptive Gaussian Densification strategy.}} We accumulate and average each Gaussian point’s gradient over training steps. Then, at each timestep \(t\), we independently apply densification or pruning based on its averaged gradient. For example, when a Minion’s mouth opens at \(F_t\), we densify that region; when it closes at \(F_T\), we prune it.}
\label{fig:densifi}
\end{figure}

3D Gaussian Splatting employs point densification to adjust Gaussian density for accurate 3D reconstruction, while 4D Gaussian Splatting~\cite{wu20244d} uses a fixed view-space gradient threshold. STAG4D~\cite{zeng2024stag4d} adds an adaptive threshold but averages gradients over all frames. As a result, all frames retain the same number of Gaussian points after these operations. We observe that the same number for all frames is suboptimal for dynamic scenes. For example, as in Fig. \ref{fig:densifi}, capturing the details of Minions' mouth requires larger numbers of Gaussian points when the mouth is suddenly open. To address this, we propose per-frame adaptive densification and pruning, dynamically tuning Gaussian counts to better model rapid spatial changes.

\noindent
\textbf{Per-Frame Adaptive Gaussian Densification.} As shown in Fig. \ref{fig:densifi}, our method analyzes the accumulated gradient $\mathcal{G}(p)$ for each Gaussian point $p$ over time, following a \emph{log-normal} distribution throughout training. To ensure adaptive selection, we define the per-frame densification threshold $\tau_t$ as:
\begin{equation}
    \tau_t = \operatorname{Quantile}_{(1-\lambda)} \big( \{\mathcal{G}(p) \mid p \in \hat{\mathcal{F}}_{t} \} \big),
\end{equation}
where \(\operatorname{Quantile}_{(1-\lambda)}(\cdot)\) represents the \((1 - \lambda)\)-quantile of the accumulated gradients over all Gaussian points and $\hat{\mathcal{F}}_{t}$ represents the set of all Gaussian points in the frame $t$. A Gaussian point \(p\) is densified only if \(\mathcal{G}(p)\ge\tau_t\).

\begin{figure*}[t!] 
\begin{center}
 \includegraphics[clip=true, trim=0in 0in 0in 0in, width=0.9\linewidth]{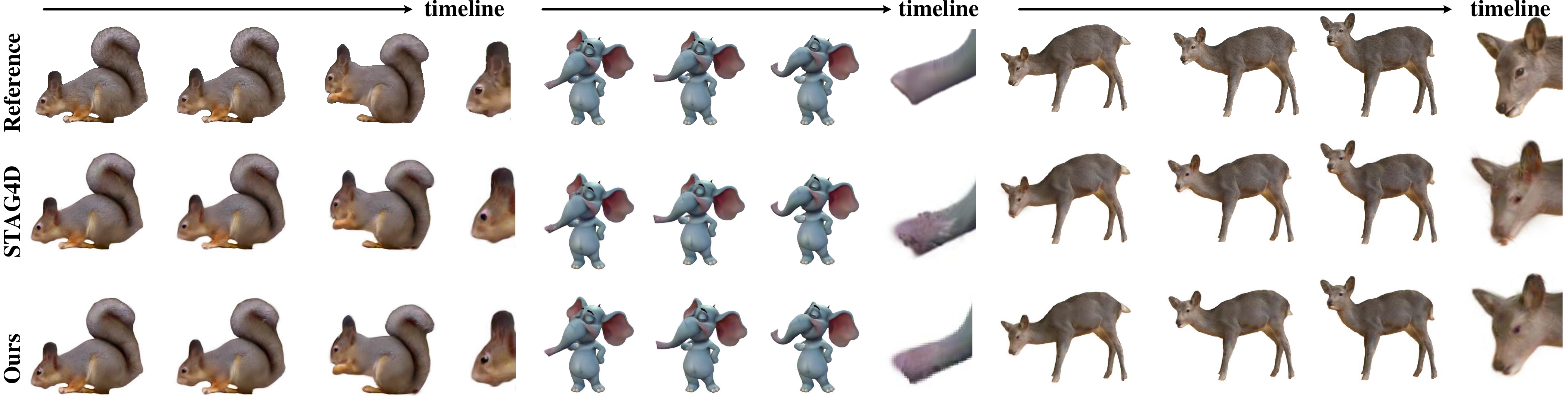} 
\end{center}
  \caption{{\textbf{Qualitative comparisons on video-to-4D generation.}} Compared with the recent SOTA method STAG4D \cite{zeng2024stag4d}, our method delivers higher-quality results in dynamic regions such as squirrel and deer heads or an elephant’s trunk.}
\label{fig:video4d}
\end{figure*}

\begin{table*}[t]
  \centering
    \centering
          \centering
          
          \begin{tabular}{l|c|cccc}
            \hline\toprule
             Methods & Reference& FID-VID $\downarrow$ & FVD $\downarrow$& CLIP $\uparrow$ & LPIPS $\downarrow$  \\
            \midrule
            DG4D~\cite{ren2023dreamgaussian4d}&arXiv'23&73 & 856& 0.88 &0.14  \\
            Consistent4D~\cite{jiangconsistent4d}& ICLR'24 & / &1134 &0.88 &0.13 \\
            4DGen~\cite{yin20234dgen} &arXiv'24& 72& /& 0.89& 0.13 \\
            SC4D~\cite{wu2024sc4d}&ECCV'24 &/ & 880& 0.90 &0.14 \\
            STAG4D~\cite{zeng2024stag4d} &ECCV'24 &53 & 992& 0.91 &0.13 \\
            4Diffusion~\cite{zhang20244diffusion} &NeurIPS'24 & /& /& 0.88&  0.17 \\
            MVTokenFlow~\cite{huangmvtokenflow}&ICLR'25 & / & 846& 0.91 & \bf0.12 \\
            4DSTR (Ours) &---& \bf45 & \bf795& \bf0.92 &\bf0.12  \\
          \bottomrule\hline
          \end{tabular}
    \caption{Quantitative comparisons with SOTA methods on the video-to-4D generation task. The best results are in \textbf{bold}.}
    \label{tab:comparison}
  \hfill
 
\end{table*}

\noindent
\textbf{Per-Frame Gaussian Pruning Strategy.} To maintain a balanced distribution of Gaussian points, we prune points based on opacity, screen-space size, and world-space scaling constraints~\cite{wu20244d,zeng2024stag4d}. Specifically, a Gaussian point \( p \) is removed if its opacity \( \sigma \) falls below a predefined threshold \( \tau_{o} \). Furthermore, points are pruned if their world-space scaling exceeds a maximum threshold \( s_{\max} \) or falls below a minimum threshold \( s_{\min} \), ensuring that only points contributing meaningfully to the reconstruction are retained as:
\begin{equation}
    \hat{\mathcal{F}}_{t}' = \{ p \in \hat{\mathcal{F}}_{t} \mid (\alpha(p) \geq \tau_{o}) \land (s_{\min} \leq \hat{s}_{t} \leq s_{\max}) \},
\end{equation}
where $\hat{\mathcal{F}}_{t}'$ indicates the Gaussian attributes that are spatially rectified after the per-frame adaptive Gaussian pruning. 
 By dynamically adjusting the per-frame densification threshold and selecting the top $\lambda$ of Gaussian points with the highest gradients, our method continuously refines the point distribution in response to scene complexity. As shown in Fig. \ref{fig:adaptive1}, our per-frame adaptive densification and pruning strategy effectively enhances the quality and robustness of 4D Gaussian representations, significantly outperforming the same number for all frames approaches in rapidly changing dynamic scenes (see Table~\ref{tab:ablation}).
 
\noindent
\textbf{Gaussian Correspondence Alignment.} Per-Frame Gaussian Densification and Pruning may disrupt inter-frame correspondence of Gaussian points, but the Temporal Rectification relies on the temporal correspondence of Gaussian points across frames. To tackle the problem, we design a per-frame index to explicitly indicate and associate each densified or pruned Gaussian point with its corresponding frame. This process ensures temporal alignment between Gaussian points in the memory buffer after the Per-Frame Gaussian Densification and Pruning.

\subsection{Loss Function}
\label{subsec:training_objectives}
Given a monocular reference video, we use Zero123++~\cite{shi2023zero123++} to render six anchor views \(\{I^i_t\}_{i=1}^6\) plus a reference \(I^{\mathrm{ref}}_t\), enhanced with temporally consistent diffusion from STAG4D~\cite{zeng2024stag4d}. We then optimize 4D Gaussians via multi-view SDS:
\begin{equation}
    \mathcal{L}_{\text{MVSDS}}
    \;=\;
    \lambda_1 \,\mathcal{L}_{\text{SDS}}\bigl(\phi,\,I^i_t\bigr)
    \;+\;
    \lambda_2 \,\mathcal{L}_{\text{SDS}}\bigl(\phi,\,I^{\text{ref}}_t\bigr),
\end{equation}
where the index $i$ is chosen based on the closest viewpoint match between the rendered images and the reference, and $\lambda_1,\lambda_2$ are weighting parameters.

Following the setup in \cite{zeng2024stag4d}, we incorporate the reference image to calculate a reconstruction term $\mathcal{L}_{\text{rec}}$ and a foreground mask term $\mathcal{L}_{\text{mask}}$. Hence, our total objective is:
\begin{equation}
    \label{eq:final_loss}
    \mathcal{L}
    \;=\;
    \mathcal{L}_{\text{MVSDS}}
    \;+\;
    \lambda_3 \,\mathcal{L}_{\text{rec}}
    \;+\;
    \lambda_4 \,\mathcal{L}_{\text{mask}},
\end{equation}
where $\lambda_3$ and $\lambda_4$ are additional coefficients. During training, we first apply \(\mathcal{L}\) to a static frame to obtain a canonical 3D Gaussian, then use anchor and reference views to learn dynamic 4D Gaussians.

Since 4DSTR learns temporal variations directly, per-frame losses alone are insufficient. Inspired by MOTR~\cite{zeng2022motr}, we define a collective average loss (CAL), which aggregates losses over a sub-clip of \(T_s\) frames as \(\mathcal{L}_{\mathrm{CAL}} = \frac{1}{T_s}\sum_{t=1}^{T_s}\mathcal{L}_{t}\). where \(\mathcal{L}_t\) denotes the total loss for frame \(t\), computed according to Eq.~\eqref{eq:final_loss}.

\section{Experiment}

\subsection{Datasets and Metrics}
\textbf{Datasets.} In terms of the video-to-4D task, we follow Consistent4D \cite{jiangconsistent4d}, leveraging multi-view videos with 7 dynamic objects to conduct the quantitative comparisons. Moreover, we also supplement the data from the online video resources created in STAG4D \cite{zeng2024stag4d}. 

\noindent\textbf{Evaluation Metrics.} We adopt four metrics \cite{jiangconsistent4d,zeng2024stag4d} to evaluate the performance of our models: CLIP and LPIPS for per-frame semantic and reconstruction quality, and FID-VID and FVD for video-level temporal coherence and multi-frame consistency.

\begin{table}[t]
    \centering
    \resizebox{1\linewidth}{!}{
    \begin{tabular}{@{}cc|cccc@{}}
      \hline\toprule
       \multicolumn{2}{c|}{Rectification} & \multicolumn{4}{c}{Evaluation Metrics} \\
      Temporal & Spatial & FID-VID $\downarrow$ & FVD $\downarrow$& CLIP $\uparrow$ & LPIPS $\downarrow$ \\
      \midrule
         & &74.24 & 1049.32 &0.902 & 0.135  \\
        & \checkmark&55.32 & 970.65 &0.910 & 0.128 \\
        \checkmark &&52.21 &850.32 & 0.912 &0.126  \\
       \checkmark& \checkmark&\textbf{45.31} & \textbf{795.21} &\textbf{0.918} &\textbf{0.121} \\
      \bottomrule\hline
    \end{tabular}}
    \caption{Ablation study on effectiveness of temporal and spatial rectification methods for video-to-4D generation.}
    \label{tab:ablation}
\end{table}

\begin{table}[t]
    \centering
    \resizebox{1\columnwidth}{!}{
    \begin{tabular}{lcccc}
      \hline\toprule
      Model         & FID-VID ↓   & FVD ↓     & CLIP ↑   & LPIPS ↓ \\
      \midrule
     
      STAG4D        & 76.00          & 1035.00         & 0.903          & 0.146          \\
       4DSTR (Ours) & \textbf{43.72} & \textbf{733.24} & \textbf{0.921} & \textbf{0.125} \\
      \bottomrule\hline
    \end{tabular}}
    \caption{Performance on extended video sequences}
    \label{tab:ablation_60}
\end{table}

\subsection{Implementation Details}
\label{subsec:setup}
The deformation fields are parameterized by MLPs with 64 hidden layers of 32 units, and the temporal model uses 32 units. During training, the learning rate decays from $1.6\times10^{-4}$ to $1.6\times10^{-6}$. For per-frame adaptive densification and pruning, we densify the top $\lambda=2.5\%$ points by accumulated gradient and prune points with opacity below $\tau_{o}=0.01$ or scale outside $[s_{\min}=0.001, s_{\max}=0.1]$~\cite{zeng2024stag4d}. We use temporal buffers of length $T=10$ and $T_s=4$, SDS weights $\lambda_1,\lambda_2=1$, and allocate $\lambda_3=2\times10^4$, $\lambda_4=5\times10^3$ to reconstruction and mask losses. The model renders at 80 FPS, with the temporal module adding approximately 0.1 M parameters and 0.23 GiB memory. All experiments run on a single RTX 4090.

\subsection{Qualitative Results}
We compare to  the state-of-the-art  method STAG4D \cite{zeng2024stag4d}, which leverages spatial–temporal anchors. As shown in Fig.~\ref{fig:vis}, our rectification yields better spatial-temporal consistency in dynamic regions. Fig.~\ref{fig:video4d} further shows our reconstructed videos offer improved rendering quality, particularly in high-dynamic regions.


\subsection{Quantitative Results and Comparison}

We quantitatively compare 4DSTR with DG4D~\cite{ren2023dreamgaussian4d}, Consistent4D~\cite{jiangconsistent4d}, 4DGen~\cite{yin20234dgen}, SC4D~\cite{wu2024sc4d}, and SOTA methods STAG4D, 4Diffusion~\cite{zhang20244diffusion}, MVTokenFlow~\cite{huangmvtokenflow} (Table~\ref{tab:comparison}). 4DSTR leads on all four metrics. It surpasses MVTokenFlow and STAG4D on image reconstruction. Due to our designed spatial-temporal rectification methods, 4DSTR has much more potential advantages in video-based metrics, with a 15.1\% FID-VID reduction and a 19.9\% FVD reduction compared to STAG4D. This demonstrates the effectiveness of our temporal correlation in enhancing spatial-temporal consistency.

\subsection{Ablation Studies}

\begin{figure}[t!] 
\begin{center}
 \includegraphics[clip=true, trim=0in 0in 0in 0in, width=0.7\linewidth]{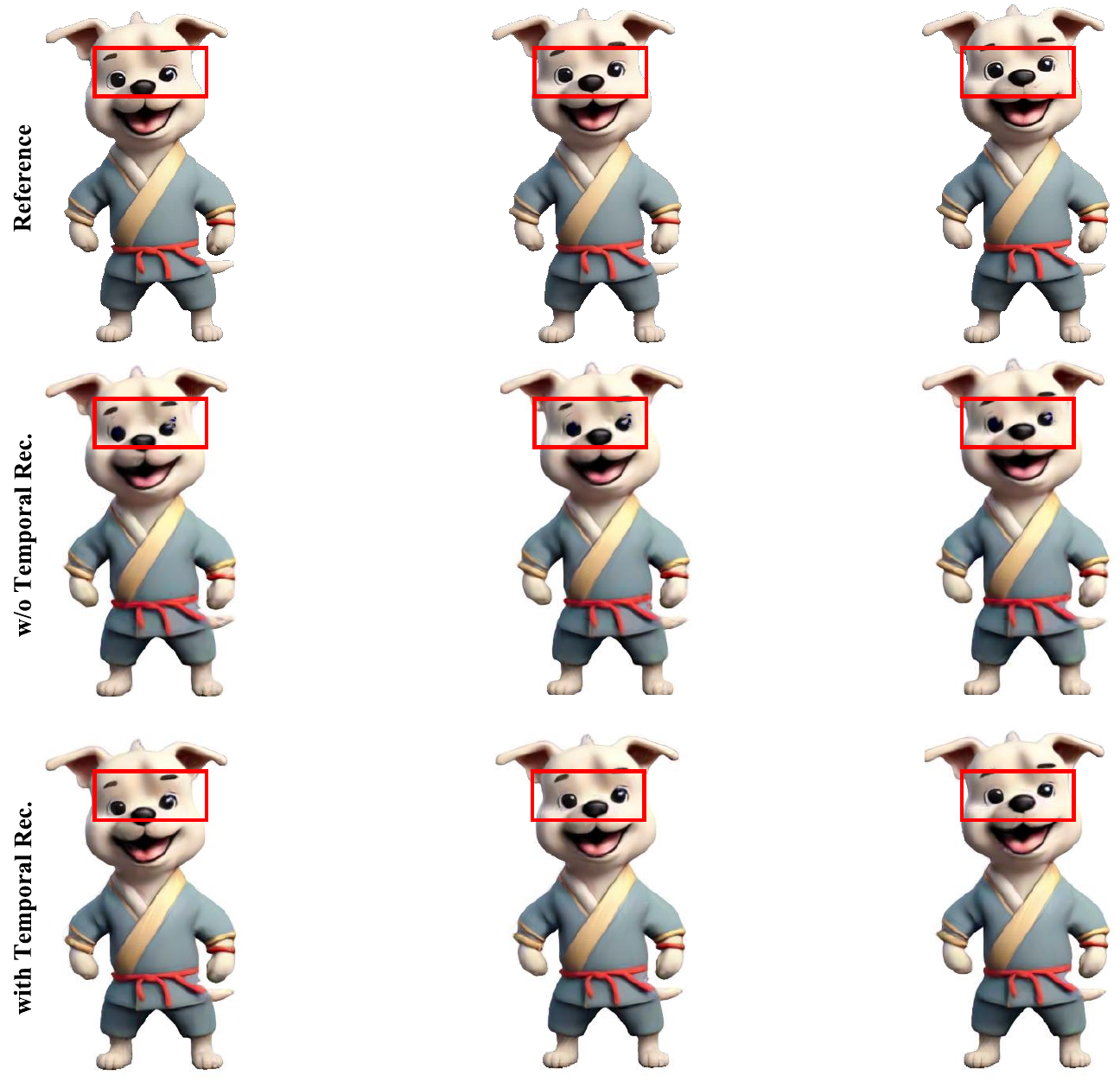} 
\end{center}
  \caption{\textbf{Ablation on temporal rectification.} Without our temporal correlation module, the temporal consistency of generated 4D sequences is largely undermined.}
\label{fig:ablation_tem}
\end{figure}

\begin{figure}[t!] 
\begin{center}

 \includegraphics[clip=true, trim=0in 0in 0in 0in, width=0.8\linewidth]{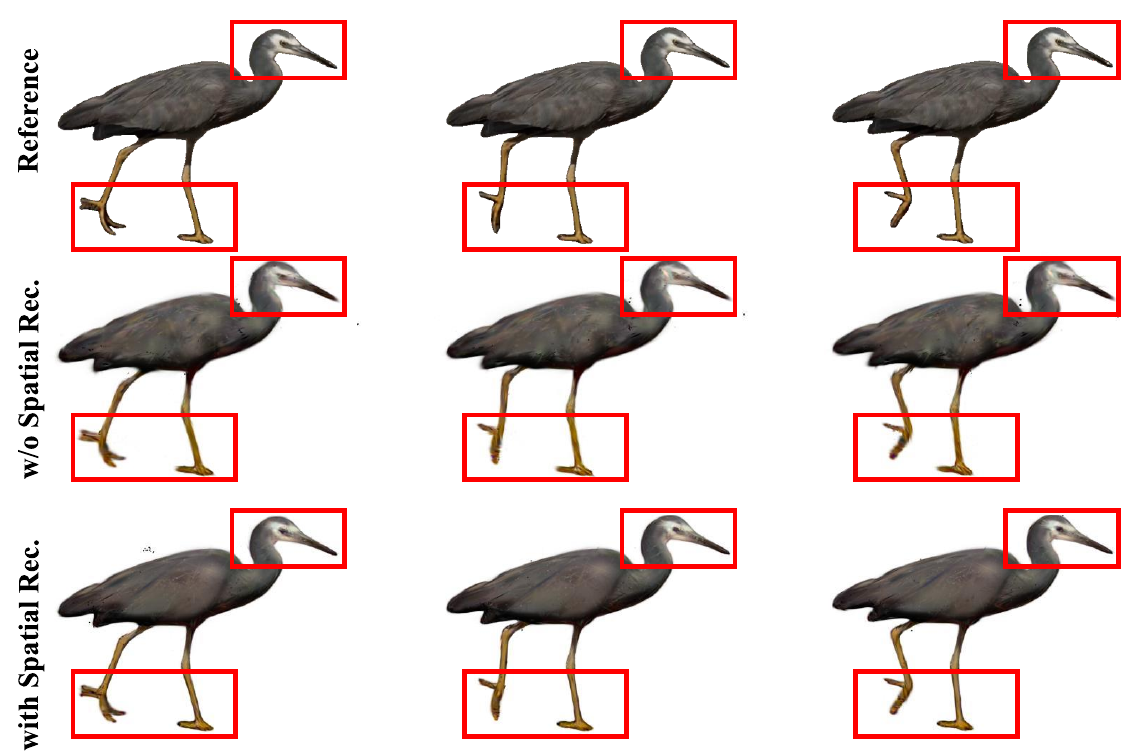} 
\end{center}
  \caption{{\textbf{Ablation on spatial rectification.}} Without our designed spatial Gaussian densification and pruning module, the reconstruction quality on high-dynamic regions is poor. We highlight differences with red rectangles.}
\label{fig:ablation_spa}
\end{figure}

\noindent\textbf{Significance of Temporal and Spatial Rectification.}
We ablate the temporal rectification and spatial densification modules in Table~\ref{tab:ablation}. Removing temporal correlation increases FID-VID and FVD by 22.1\%, highlighting its role in maintaining temporal consistency (Fig.~\ref{fig:ablation_tem}). Spatial rectification with adaptive Gaussian densification and pruning is likewise crucial for high-quality rendering of fast-moving objects (Fig.~\ref{fig:ablation_spa}).

\noindent\textbf{Longer Sequence Generation Results.}  
To validate robustness over longer horizons, we construct a 60-frame test set following Consistent4D in Table~\ref{tab:ablation_60}. While STAG4D’s performance degrades sharply on 60-frame inputs, 4DSTR further reduces FID-VID and FVD, confirming that our spatial-temporal rectification mechanism scales effectively to extended sequences, preserving spatial-temporal consistency.

\begin{figure*}[!t]
  \centering
  \includegraphics[clip, trim=0in 0in 0in 0in, width=0.8\linewidth]{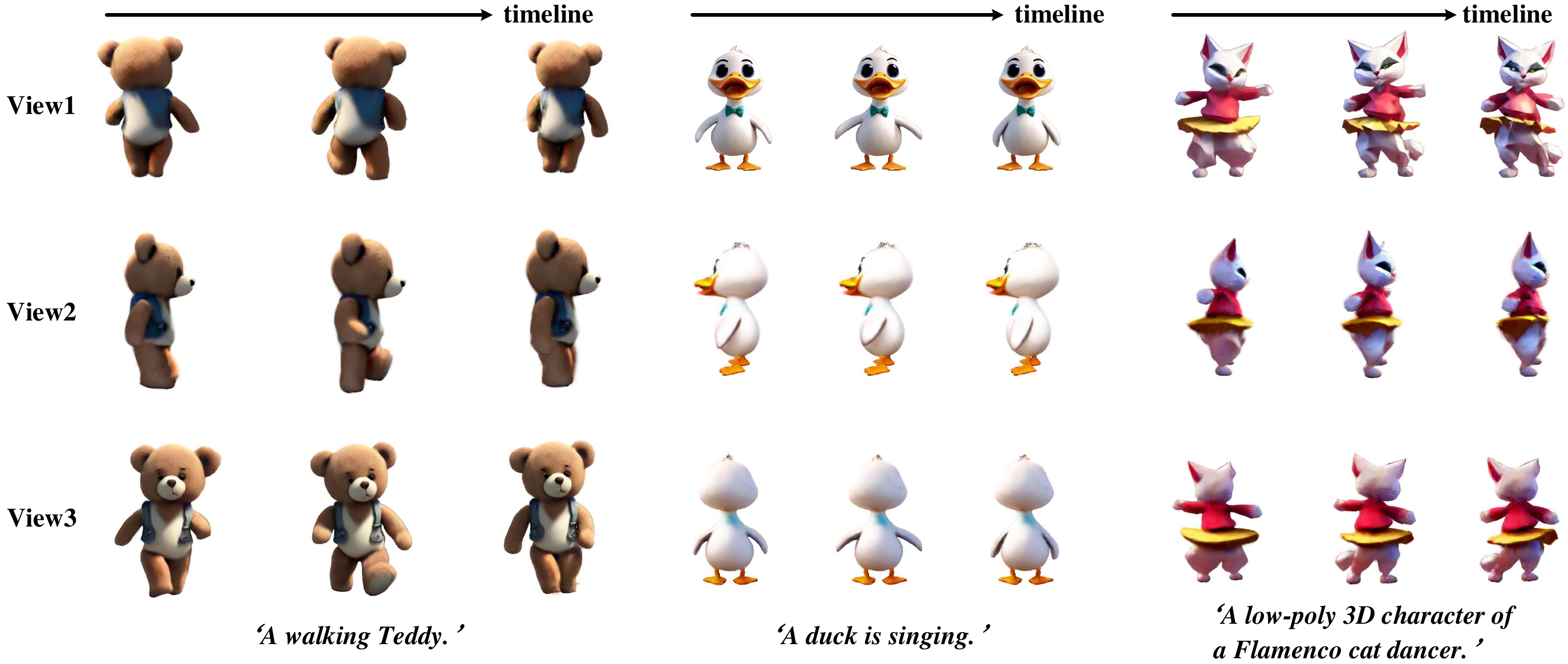}
  \caption{\textbf{Qualitative comparisons on text-to-4D generation.} Our method can also support text information as the guidance, generating consistent and high-quality 4D sequences which can be observed from diverse views.}
  \label{fig:text4d}
\end{figure*}

\noindent
\textbf{Various Temporal Encoding Methods.} We adopt a Mamba-based interaction for temporal encoding of Gaussian scales and rotations, leveraging its linear-complexity modeling of long-range dependencies. Table~\ref{tab:temporal_attn} shows that Mamba outperforms GRU~\cite{cho2014learning}  and attention~\cite{vaswani2017attention} backbones, achieving up to a 5.01 reduction in FID-VID and running at 80\,FPS, which led us to choose it for our temporal correlation module.

\begin{table}[t]
    \centering
    \resizebox{\linewidth}{!}{%
    \begin{tabular}{@{}lccccc@{}}
      \hline\toprule
      & \multicolumn{4}{c}{Evaluation Metrics} & \multirow{2}{*}{Latency (fps) $\uparrow$} \\
      \cmidrule(lr){2-5}
      \multirow{-2}{*}{Methods} 
      & FID-VID $\downarrow$ 
      & FVD $\downarrow$ 
      & CLIP $\uparrow$ 
      & LPIPS $\downarrow$ 
      &  \\ 
      \midrule
      GRU      
        & 50.32   & 821.13 & 0.913 & 0.127 
        & 68 \\
      Attention 
        & 54.23   & 812.36 & 0.914 & 0.125 
        & 72 \\
      Mamba                      
        & \textbf{45.31} & \textbf{795.21} & \textbf{0.918} & \textbf{0.121} 
        & \textbf{80} \\
      \bottomrule\hline
    \end{tabular}%
    }
    \caption{Ablation study on temporal correlation methods.}
    \label{tab:temporal_attn}
\end{table}

\begin{table}[t]
    \centering

    \resizebox{1\linewidth}{!}{
    \begin{tabular}{@{}lcccc@{}}
      \hline\toprule
      & \multicolumn{4}{c}{Evaluation Metrics} \\
       \multirow{-2}{*}{Temporal lengths}& FID-VID $\downarrow$ & FVD $\downarrow$& CLIP $\uparrow$ & LPIPS $\downarrow$  \\
      \midrule
      T=2 & 57.32 &855.13 & 0.908 & 0.132 \\
      T=5 & 53.21 &823.32&0.912 & 0.127 \\
      T=10 &45.31 & \textbf{795.21}&\textbf{0.918} &0.121  \\
      T=15 & \textbf{43.54}&  804.32&0.917 & \textbf{0.120}\\
      \bottomrule\hline
    \end{tabular}}
    \caption{Ablation study on different temporal length.}
    \label{tab:temporal_length}
\end{table}

\noindent\textbf{Varying Temporal Window Sizes.}  
Although rapid variations are local, adjacent‐frame models often fail under occlusions, appearance shifts, and large displacements. Extending the temporal window provides additional context to resolve abrupt motions and preserve spatial–temporal coherence. To validate this, we vary the window size \(T\) and report results in Table~\ref{tab:temporal_length}, which show that increasing \(T\) from 2 to 5 reduces FID-VID by 14.0\% and FVD by 15.2\%, and extending to \(T=10\) yields further reductions of 13.6\% and 17.8\%, respectively. For \(T>10\), all metrics stabilize, confirming that a 10-frame window suffices to capture rapid variations.

\begin{table}[t!]
    \centering
    \resizebox{1\linewidth}{!}{
    \begin{tabular}{@{}lccc@{}}
      \hline\toprule
      Methods& Vis. & Cons.& Align.  \\
      \midrule
Consistent4D \cite{jiangconsistent4d}& 13.3\% & 20.0\% &16.7\% \\
      STAG4D \cite{zeng2024stag4d}&33.3\% & 30.0\% & 36.7\%\\
      Ours &\bf53.3\%& \bf50.0\% & \bf46.7\%\\
      \bottomrule\hline
    \end{tabular}}
    \caption{User study for the text-to-4D generation task. }
    \label{tab:user}
\end{table}

\subsection{Results of Text-to-4D Generation}
Following prior works \cite{zeng2024stag4d}, our pipeline also supports text-to-4D generation by using SDXL \cite{podell2023sdxl} for image synthesis and SVD \cite{blattmann2023stable} to create short videos, then applying the video-to-4D pipeline above. We use the same data settings as STAG4D for fair comparison and conduct a user study \cite{huangmvtokenflow} evaluating the performance of text-to-4D generation task. Fig.~\ref{fig:text4d} shows diverse 4D samples with plausible dynamics and spatio-temporal consistency across modalities. Following \cite{zeng2024stag4d}, the user study covers 14 test scenarios with 30 evaluators rating visual quality (Vis.), temporal consistency (Cons.), and alignment with the input text (Align.); in Table~\ref{tab:user}, our method achieves the highest scores on all metrics, demonstrating the superiority of 4DSTR on the text-to-4D generation task.

\section{Conclusion}

In this paper, we introduce 4DSTR, a 4D generation network with spatial–temporal rectification. Our approach correlates deformable 4D Gaussian points across multiple frames, ensuring a consistent Gaussian representation in each frame. 
To handle rapid temporal changes, we introduce an adaptive spatial densification and pruning strategy, dynamically adding or removing Gaussian points based on long-range dependencies. 
Extensive experiments show 4DSTR sets a new SOTA in video-to-4D generation on reconstruction quality, temporal coherence, and dynamic adaptability. Furthermore, our designed pipeline can also support high-quality text-to-4D generation task when combined with existing text-to-image generators.

\section{Acknowledgments}

This work was supported by the EU HORIZON-CL4-2023-HUMAN-01-CNECT XTREME (grant no.101136006), and the Sectorplan Beta-II of the Netherlands.

\bibliography{aaai2026}

\end{document}